\pdfoutput=1

\documentclass[11pt]{article}

\usepackage{EMNLP2022}

\usepackage{times}
\usepackage{latexsym}

\usepackage[T1]{fontenc}

\usepackage[utf8]{inputenc}

\usepackage{microtype}

\usepackage{inconsolata}

\usepackage{graphicx}
\usepackage{paralist}
\usepackage{url}
\usepackage{caption}
\usepackage{xspace}
\usepackage{amsfonts,amsmath}
\usepackage{booktabs}
\usepackage{setspace}
\usepackage{multirow}
\usepackage{multicol}
\usepackage{subfigure}
\usepackage{hyperref}
\usepackage{amssymb}
\usepackage{pifont}
\usepackage{color, colortbl}
\usepackage{bbm}
\usepackage{wrapfig}
\usepackage{lipsum}
\usepackage{enumitem}
\usepackage{paralist, tabularx}
\usepackage{balance}

\title{A Unified Encoder-Decoder Framework with Entity Memory}

\author{{\bf Zhihan Zhang$^1$, Wenhao Yu$^1$, Chenguang Zhu$^2$, Meng Jiang$^1$} \\
$^1$University of Notre Dame, Notre Dame, IN, USA \\
$^2$Microsoft Cognitive Services Research, Redmond, WA, USA \\
\normalsize{ $^1${\tt \{zzhang23, wyu1, mjiang2\}@nd.edu}; $^2${\tt chezhu@microsoft.com}}
}

\begin{document}
\maketitle

\begin{abstract}

Entities, as important carriers of real-world knowledge, play a key role in many NLP tasks.
We focus on incorporating entity knowledge into an encoder-decoder framework for informative text generation. Existing approaches tried to index, retrieve, and read external documents as evidence, but they suffered from a large computational overhead. 
In this work, we propose an \textbf{E}ncoder-\textbf{D}ecoder framework with an entity \textbf{Mem}ory, namely EDMem. The entity knowledge is stored in the memory as latent representations, and the memory is pre-trained on Wikipedia along with encoder-decoder parameters. To precisely generate entity names, we design three decoding methods to constrain entity generation by linking entities in the memory. EDMem is a unified framework that can be used on various entity-intensive question answering and generation tasks. Extensive experimental results show that EDMem outperforms both memory-based auto-encoder models and non-memory encoder-decoder models.\footnote{Code will be available at \url{https://github.com/DM2-ND/EDMem}}

\end{abstract}
\vspace{0.1cm}

\section{Introduction}
\label{sec:introduction}
A large amount of real-world knowledge is related to entities, \textit{e.g.}, persons, nations, and events. Entity knowledge is the information describing facts and attributes related to entities. 
Many entity-intensive NLP tasks require models obtain entity knowledge to generate informative outputs, such as answering factual questions~\cite{NQ}, explaining claims~\cite{CREAK}, or making informative conversations~\cite{WoW}. 
Pre-trained encoder-decoder models can be directly applied on such entity-intensive tasks~\cite{learning_to_mask, closed_book_T5}, but their ability to store and use knowledge is still questionable~\cite{TrainTestOverlap, wang2021generative}. A popular approach to incorporate knowledge into the generation process is retrieving evidence documents from external sources~\cite{RAG, FiD, UniKQA, kenlg_survey}. However, they suffer from significant computational overheads in indexing, retrieving, and reading a large number of extra documents~\cite{DensePhrases, MentionMemory}. Therefore, it is important to give encoder-decoder models access to entity knowledge without sacrificing too much efficiency.

\begin{figure}
    \centering
    \includegraphics[width=0.48\textwidth]{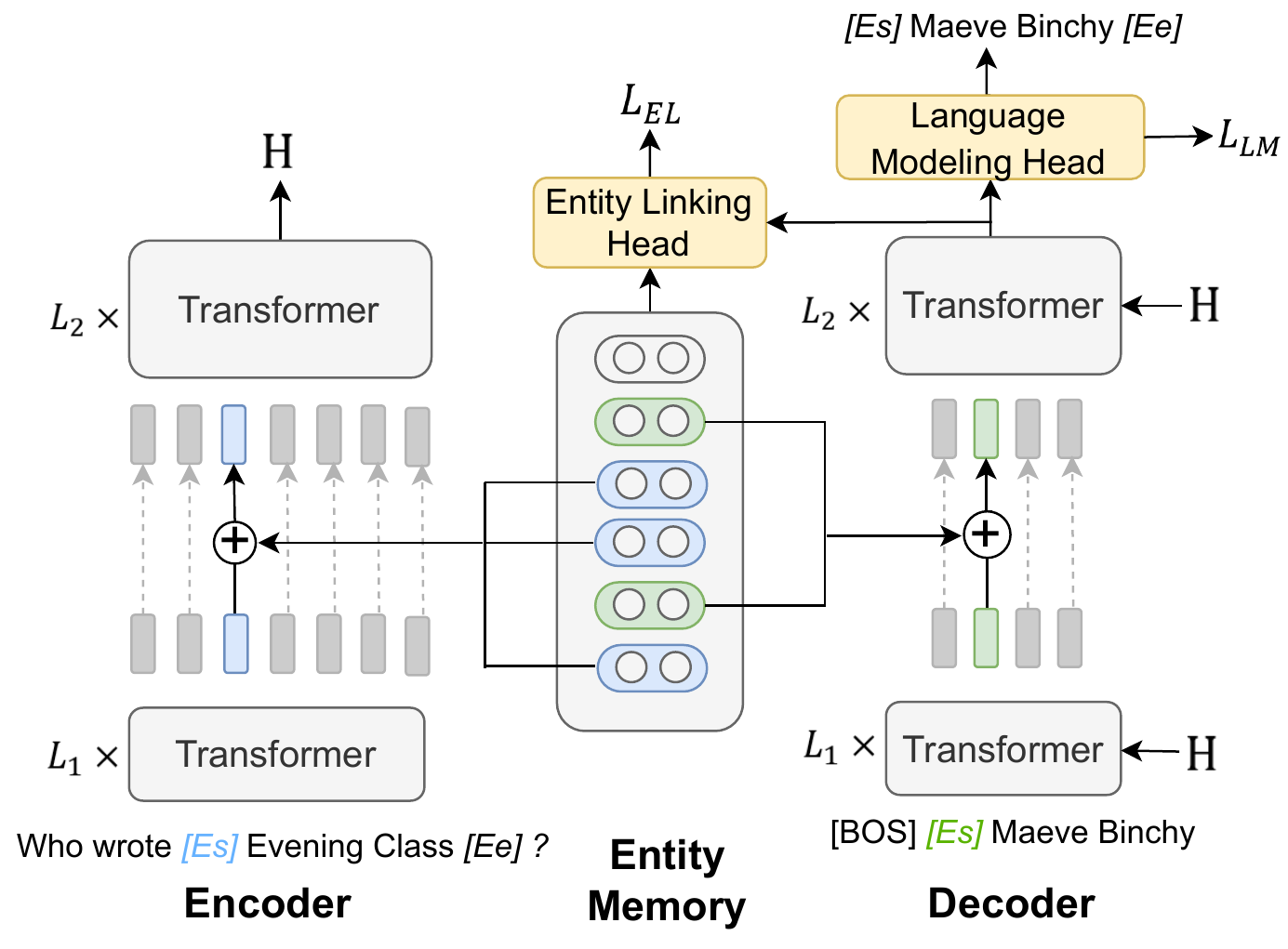}
    \caption{An overview of the EDMem framework. $\mathbf{H}$ denotes the final hidden states of the encoder.}
    \label{fig:model}
    \vspace{-0.2cm}
\end{figure}

Recently it has been proposed to use
an in-model memory
to augment auto-encoder models with entity knowledge on entity linking tasks~\cite{EaE, FaE, OPQL}.
The entity memory stores entity knowledge as dense vectors which can be directly incorporated into the hidden states of Transformer models~\cite{transformer}, with no need to encode extra text. However, the auto-encoder framework in previous approaches can only select entities from a pre-defined entity vocabulary. Hence, they are not able to give an entity outside the vocabulary, nor to \textit{generate} answers or text beyond a single entity.



In this paper, we propose a novel \textbf{E}ncoder-\textbf{D}ecoder framework with an entity \textbf{Mem}ory (EDMem), as shown in Figure~\ref{fig:model}.
EDMem is a unified framework on various entity-intensive QA and  generation tasks, in which we train an entity memory for efficient knowledge incorporation.
First, EDMem is pre-trained on Wikipedia documents, where it learns entity embeddings in the memory along with an encoder-decoder model. EDMem learns to select relevant entities from the memory via an entity linking objective, and learns to generate answers using entity knowledge via a language modeling objective. Second, to precisely generate entity names, we design three decoding methods that utilize the entity linking ability of EDMem in its generation process, when we fine-tune it on downstream tasks. These include (1) free-form: left-to-right generation with entity identifiers; (2) static entity linking: first select entities by entity linking, build prefix trees for the selected entities, and then perform constrained entity generation using the trees; (3) dynamic entity linking: select entities on-the-fly for constrained entity generation.
We conduct experiments on two popular testbeds of entity knowledge: open-domain QA and entity-intensive generation. With the incorporation of entity knowledge, EDMem outperforms non-memory encoder-decoder models on both tasks, and it retains the efficiency advantage of closed-book (\textit{i.e.}, non-retrieval) models. Compared to memory-based auto-encoders, EDMem achieves both higher overall accuracy (+9\%) and better entity precision (+8\%) on open-domain QA datasets, and it generates high-quality text from the memory-supported decoder on generation datasets when auto-encoders fail to do so. To summarize, EDMem is the first knowledge-augmented closed-book framework to perform both tasks in a \emph{unified} manner.

\section{Related Work}
\label{sec:related}


\paragraph{Closed-Book Models} Closed-book models are pre-trained models that store knowledge in their own parameters.
For example, 
COMET~\cite{COMET} fine-tuned GPT2~\cite{GPT2} to construct knowledge graphs by generating commonsense triples.
Recently, 
fine-tuned BART~\cite{BART} or T5~\cite{T5} models are proved to be competitive on open-domain QA~\cite{learning_to_mask, closed_book_T5}. 
Therefore, closed-book models are able to memorize some entity knowledge after pre-trained on massive data. However, studies showed that closed-book models just recalled similar inputs and answers in their pre-training corpus~\cite{wang2021generative}, and their performances were behind open-book models.
\paragraph{Open-Book Models} Open-book models first retrieve evidence documents from external corpora and read these documents to predict an answer~\cite{DrQA}.
REALM~\cite{REALM} proposed a self-supervised approach to pre-train a retriever-reader model.
DPR~\cite{DPR} devised a contrastive objective to train a dense bi-encoder retriever on open-domain QA.
Subsequent approaches combined DPR with a generative objective to build large, powerful models on open-domain QA and generation tasks~\cite{RAG, FiD, sachan2021end, KG-FiD}. 
However, open-book models have to process the raw text of all retrieved documents, which leads to extremely long inference time. Besides, additional overheads are brought by loading the document index and retrieving evidence documents for each example.

\paragraph{Entity Memory} 
EaE~\cite{EaE} was the first to pre-train an entity memory with an auto-encoder framework to perform entity prediction on open-domain QA. 
FILM~\cite{FaE} followed EaE and added a fact memory containing representations of Wikidata triples. To better encode relational knowledge, OPQL~\cite{OPQL} learned latent relational representations for arbitrary entity pairs. Recent work focused on learning a huge mention-level memory (\textasciitilde150M entries) with extensive pre-training~\cite{MentionMemory} or leveraging the entity memory in domain adaptive training~\cite{KALA}. These models are all based on an auto-encoder framework. Thus, they are able to predict entities IDs but would fail to generate any non-entity answers or sentences. 
There is a preprint paper contemporaneous to our work which trained a memory with an encoder-decoder model~\cite{QAMemory}. However, it used QA pairs as memory entries instead of entities, limiting its application to QA tasks. Besides, their memory is much heavier (60M entries) than ours (1M).

\section{Proposed Framework}
\label{sec:method}
Suppose we have a pre-defined vocabulary of $N$ {entities} $\mathcal{E}=\{e_1,\dots,e_N\}$. A {mention} is the actual tokens in context which refer to an entity. The set of all mentions in the corpus is denoted as $\mathcal{M}$. Thus, there is a global alias table $\mathcal{T}:\mathcal{E}\rightarrow2^{\mathcal{M}}$, where each entity is mapped to all its mentions. 
The input of EDMem is a sequence of tokens $\boldsymbol{x}$ 
of length $S$, and the target output is another sequence $\boldsymbol{y}=[y_1,\cdots,y_T]$ of length $T$. Both sequences contain a pre-labeled set of mentions. 
Each mention refers to an entity in $\mathcal{E}$.
We add two special tokens $[E_s]$ and $[E_e]$ to represent ``entity start'' and ``entity end'' boundaries of a mention, \textit{e.g.}, ``$[E_s]$ \textit{Brett Hart} $[E_e]$ \textit{is the president of the} $[E_s]$ \textit{United Airlines} $[E_e]$''. These special tokens come from either Wikipedia hyperlinks (in pre-training, \S\ref{sec:pre-train}) or an entity linking model (in fine-tuning, \S\ref{sec:finetune}).

\subsection{Architecture}

An overview of EDMem is presented in Figure~\ref{fig:model}. The framework has a transformer encoder, a transformer decoder, an entity memory, and two prediction heads. Both the encoder and decoder have two parts: ($L_1\times$) lower layers and ($L_2\times$) upper layers. Transformer layers in EDMem have the same architecture with BART~\cite{BART}. At the end of lower layers, EDMem is allowed to use the hidden states as a query to access the entity memory.
The knowledge representation obtained by each memory access is summed and normalized with the hidden states before performing further reasoning in upper layers.
Two prediction heads use the final hidden states of the decoder for prediction: an LM head for token prediction and an entity linking head for entity prediction (Details are in \S\ref{sec:pre-train}).
In practice, we follow EaE~\cite{EaE} to set $L_1=4$ and $L_2=8$.

\subsection{Entity Memory}
\label{sec:memory}

The entity memory contains a large embedding table, which stores the embeddings of entities in $\mathcal{E}$.
Intuitively, an entity embedding contains the contextual information around all mentions of the entity in Wikipedia documents.
During encoding and decoding, EDMem queries the entity memory whenever it encounters a mention. It recognizes mentions by identifying the $[E_s]$ token. 
EDMem takes the hidden state of the $[E_s]$ token as query to retrieve relevant knowledge from the entity memory by attending to the entity embedding table (bias terms are omitted):

\vspace{-0.65cm}
\begin{align*}
    \mathbf{h}^{ent}_{s} &= \mathbf{W}_{out}(\sum_{i=1}^{N}\alpha_i\cdot\mathbf{e}_i),  \tag{1} \label{eq:attn1}\\
    \textrm{where} \ \alpha_i &= \frac{\exp{(\mathbf{e}_i^\intercal\mathbf{W}_{in}\mathbf{h}^{low}_{s} )}}{\sum_{j=1}^{N}\exp{(\mathbf{e}_j^\intercal\mathbf{W}_{in}\mathbf{h}^{low}_{s} )}}. \tag{2}\label{eq:attn2}
\end{align*}

\noindent $\mathbf{e}_i$  is the embedding of entity $e_i$. $\mathbf{h}^{low}_{s}$ denotes the hidden state of the $[E_s]$ token (from lower encoder/decoder layers). $\mathbf{h}^{ent}_{s}$ is the aggregated entity representation, which is summed and normalized with $\mathbf{h}^{low}_{s}$ to put into upper layers. ${\mathbf{W}}_{in}$ and ${\mathbf{W}}_{out}$ are linear projection layers for dimension matching.
Following EaE, during inference, we aggregate the entity representaion of top 100 entities (sorted by $\alpha_i$) instead of attending to all $N$ entities.


\subsection{Pre-Training}
\label{sec:pre-train}

\subsubsection{Pre-Training Corpus}
We pre-train EDMem on the whole Wikipedia corpus. All documents are split into 128-token passages. 
In addition, we set a 10-token sliding window between passages to avoid an entity being split into two adjacent chunks. Such a setting yields a total of 39M passages, of which we hold out 0.5\% of them as the validation set during pre-training. We leverage Wikipedia hyperlinks as gold annotations of 249M mentions and their linked entities. Since hyperlinks do not cover all mentions in text, we heuristically label missing mentions to create more training signals for the entity memory. We use the alias table $\mathcal{T}$ to label all mentions in a Wikipedia page if they match either (1) a linked entity in the same page, or (2) the title entity of this page. This leads to a total of 468M mentions in the pre-training corpus. We collect 1M most frequently linked entities to form the entity vocabulary $\mathcal{E}$. More details can be found in Appendix~\ref{sec:appendix_pretrain}.


\subsubsection{Pre-Training Objective}
\label{sec:pretrain_obj}

Our pre-training objective is a combination of language modeling and entity linking. For language modeling objectives, we randomly corrupt parts of the input sequence and train EDMem to reconstruct the original sequence. We adopt two kinds of sequence corruption: \emph{random token masking} and \emph{salient span masking}. In random token masking, each token has a probability of $P_{rtm}$ to be replaced by a [MASK] token. Salient span masking is adapted from~\cite{REALM}, where each mention has a probability of $P_{ssm}$ that all tokens within the mention are replaced by [MASK]. Such explicit masking of whole mention names encourages EDMem to rely on the entity memory in predicting mentions, which facilitates the learning of entity embeddings. The LM head performs token prediction through a linear-softmax layer, and the LM loss is the negative log-likelihood of the target sequence: $L_{LM}=-\sum_{j=1}^{T}\textrm{log}~{P}(y_j|\boldsymbol{x}, y_{1:j-1})$.


EDMem utilizes direct supervision signals to the entity memory for entity representation learning. The entity linking loss is applied each time it queries the entity memory. Besides in the middle of the encoder and decoder, EDMem queries the memory in the entity linking head, as shown in Figure~\ref{fig:model}. The entity linking head predicts the corresponding entity using the hidden states of each mention, the same as Equation (\ref{eq:attn2}).
We use a cross-entropy loss to maximize the attention weights of the labelled entities: $L_{EL}=-\sum_{m}\textrm{log}~\alpha_i$, where $m$ is a mention in the input or output sequence that is linked to the $i$-th entity in $\mathcal{E}$. The final loss function is
$L_{LM}+\lambda_{EL}L_{EL}$, where the coefficient $\lambda_{EL}$ is a hyper-parameter. 


\begin{figure}
    \centering
    \includegraphics[width=0.5\textwidth]{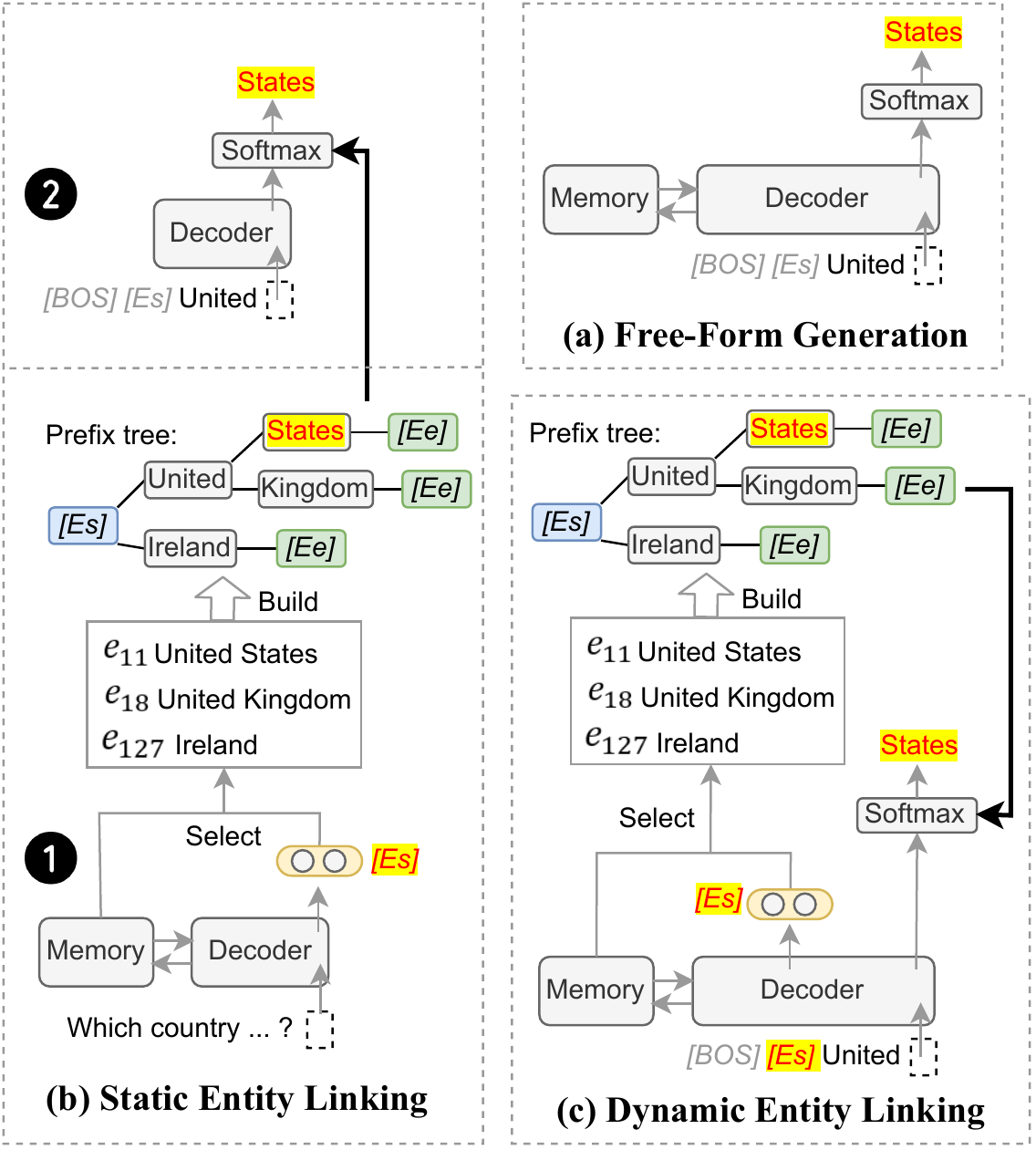}
    \caption{Three decoding methods in downstream tasks.}
    \label{fig:decoding}
\end{figure}

\subsection{Fine-Tuning}
\label{sec:finetune}


EDMem is fine-tuned on downstream tasks via an LM objective and an entity linking objective. The LM objective is to maximize the probability of the task-specific output. The entity linking objective links mentions to entities in the memory, the same as pre-training. Mention boundaries are pre-labeled using an state-of-the-art entity linking model~\cite{ELQ}.
In entity-intensive downstream tasks, the entity memory assists sequence generation by not only providing entity knowledge but also generating entity names.
Thus, we design three decoding settings to let the entity linking objective assist sequence generation. A sketch of different settings is given in Figure~\ref{fig:decoding}.

\paragraph{Free-Form Generation}

In this setting, the model generates the output sequence entirely based on the probability given by the LM head. This includes the special tokens $[E_s]$ and $[E_e]$ which indicate an access to the memory. There is no constraint on what tokens to generate between $[E_s]$ and $[E_e]$, \textit{i.e.}, the subsequence $[E_s],y_i,\cdots,y_j,[E_e]$ may not be a valid entity name in the entity vocabulary. One advantage is that the model processes the entity knowledge in a latent manner, which does not explicitly affect the probability distribution of the language model. However, this may affect the model's performance in tasks where entity names are strictly required, \textit{e.g.}, open-domain QA tasks where exact match is used as evaluation.



\paragraph{Static Entity Linking}

Static entity linking explicitly restricts the model to generate entity names for QA. Here, the decoding process is divided into two steps: \textit{entity linking} and \textit{constrained generation}.
First, given a question, the model selects one or multiple entities as references. As shown in Figure~\ref{fig:decoding}(b), the question with an appended $[E_s]$ token as a placeholder is passed into the decoder, and the entity linking head is trained to predict the entity ID of the gold answer\footnote{Training examples with non-entity answers are discarded.}.
Then we have the selected top-$k$ entities for each test question.
We restrict the generation space to the top-$k$ entities when the model is trying to generate an entity name.
To achieve this, inspired by~\cite{GENRE}, we build a prefix tree for $k$ entities for each test example. The prefix tree tells the model which tokens are allowed to generate given a prefix (\textit{i.e.}, previous generated tokens). When the model generates an $[E_s]$ token, we restrict the following generated tokens to be one of the $k$ entity names (\textit{i.e.}, one of the paths in the prefix tree). In this way, the model can either generate an entity answer (by generating $[E_s]$ and traversing the pre-built prefix tree), or generate a non-entity answer (if no $[E_s]$ token is generated). 
Readers can refer to~\cite{GENRE} for more implementation details.


\paragraph{Dynamic Entity Linking}

Static entity linking is applicable only when the downstream task can be converted into an entity linking objective. Another way to generate entities is to predict the entities \textit{on-the-fly}.
After each time the model generates an $[E_s]$ token, 
the entity linking head predicts top-$k$ entities using the hidden state of $[E_s]$ based on previous generated tokens, as shown in Figure~\ref{fig:decoding}(c).
This differs from static entity linking, where the model makes entity predictions solely dependent on the input sequence.
A prefix tree of the names of top-$k$ entities is also built on-the-fly for constrained entity generation.




\section{Experiments}
\label{sec:Experiments}
We test our EDMem framework on two testbeds of entity knowledge: open-domain QA and entity-intensive generation tasks.

\subsection{Open-Domain QA}

\subsubsection{Data}

Open-domain QA is a task where models are required to answer questions without any provided evidence. Questions are usually related to real-world facts and entities. We test EDMem on three popular datasets: Natural Questions (NQ)~\cite{NQ}, TriviaQA (TQA)~\cite{TriviaQA} and WebQuestions (WQ)~\cite{WebQ}. We follow the in-house splits introduced by~\cite{lee2019latent}. We also report on dev set of the TQA official split to compare with EaE~\cite{EaE}. We report exact match (EM) scores on these datasets.

We mainly compare with previous closed-book models (\textit{i.e.}, models without evidence retrieval), including traditional encoder-decoder models like BART~\cite{BART} and T5~\cite{T5}, and memory-based auto-encoder models like RELIC~\cite{RELIC}, EaE, and FILM~\cite{FaE}.
Besides, We pre-train two ablations of EDMem. EncMem is composed of an encoder and an entity memory, and is trained via the same objectives as EDMem. EncDec removes the entity memory from EDMem, and is trained via the same LM objectives.
We also list the performance of state-of-the-art open-book models (\textit{i.e.}, models with evidence retrieval to assist prediction) for reference, such as REALM~\cite{REALM}, RAG~\cite{RAG}, and FiD~\cite{FiD}. We test three variants of EDMem, \textit{i.e.}, free-form generation (-free), static entity linking (-stat.) and dynamic entity linking (-dyn.).

\begin{table}[t]
\centering
\setlength{\tabcolsep}{1.5mm}{
\resizebox{0.48\textwidth}{!}{
\begin{tabular}{l|cccc}
\toprule
\multirow{2}{*}{\textbf{Model}} & \multicolumn{2}{c}{\textbf{TQA}} & \textbf{NQ}    & \textbf{WQ}    \\
                                & In-House Test & Dev   & Test           & Test           \\ \midrule
\multicolumn{5}{c}{\textit{Closed-Book Models}}                                                      \\ \midrule
BART-Large*                      & 25.02           & 27.28           & 24.82          & 29.23          \\
T5-Large*                   & -           &   28.70             & 28.50          & 30.60          \\
EncDec*   & 27.54 & 30.01  & 25.96 & 29.38 \\
RELIC\dag                           & 35.70           & -              & -              & -              \\
EaE\dag                             & -               & 43.20          & -              & 39.00    \\
FILM\dag                            & 29.10           & -              & -              & -              \\
EncMem\dag                  & 41.01             & 42.00         & 25.54                 & 38.88   \\
EDMem-free               & 42.24           & 43.31          & {\underline{29.14}}    & 36.47          \\
EDMem-stat.              & \textbf{46.19}  & \textbf{47.23} & \textbf{30.19} & \textbf{41.44} \\
EDMem-dyn.        & {\underline{43.82}}     & {\underline{44.44}}    & 27.70          & \underline{39.52}          \\ \midrule
\multicolumn{5}{c}{\textit{Open-Book Models}}                                                        \\ \midrule
REALM                           & -               & -              & 40.40          & 40.70          \\
RAG                             & 56.80  & -              & 44.50 & 45.20 \\
FiD                             & 67.60  & -              & 51.40 & 47.64 \\\bottomrule
\end{tabular}}}
\caption{Exact match scores on open-domain QA datasets. \textbf{Bold} scores and \underline{underlined} scores are the best and second best results among closed-book models. (*traditional encoder-decoder models, \dag memory-based auto-encoder models)}
\label{tab:openqa}
\vspace{-0.4cm}
\end{table}

\begin{table*}[t]
\centering
\setlength{\tabcolsep}{1.3mm}{
\resizebox{\textwidth}{!}{
\begin{tabular}{l|cccccc|ccc|ccc}
\toprule
\multicolumn{1}{l|}{\multirow{3}{*}{\textbf{Model}}} & \multicolumn{6}{c|}{\textbf{TQA}}                                                                            & \multicolumn{3}{c|}{\textbf{NQ}}                           & \multicolumn{3}{c}{\textbf{WQ}}                          \\
\multicolumn{1}{c|}{}                       & \multicolumn{3}{c}{In-House Test}               & \multicolumn{3}{c|}{Official Dev}                 & \multicolumn{3}{c|}{Test}                         & \multicolumn{3}{c}{Test}                        \\
                                           & Total          & Ent.       & Non-Ent.  & Total          & Ent.       & Non-Ent.   & Total          & Ent.       & Non-Ent.   & Total          & Ent.       & Non-Ent.  \\ \midrule
BART                                       & 25.02          & 27.92          & 9.31          & 27.28            &  30.52           & 9.70           & 24.82          & 28.54          & 15.95          & 29.23          & 32.28          & 5.24          \\
EaE                                        & -              & -              & -             & 43.20          & 51.43          & 0.00           & -              & -              & -              & 39.00          & 42.86          & 0.00          \\
EncMem                                    & 41.01            & 48.58          & 0.00          & 42.00          & 49.74          & 0.00           & 25.54          &  36.24 & 0.00           &   38.88   &   43.82  & 0.00          \\
EDMem-free                                  & 42.24          & 48.27          & \textbf{9.59} & 43.31          & 49.44          & \textbf{10.10} & 29.14          & 34.32          & \textbf{16.19} & 36.47          & 40.16          & \textbf{7.42} \\
EDMem-stat.                                & \textbf{46.19} & \textbf{53.82} & 4.88          & \textbf{47.23} & \textbf{54.86} & 5.85           & \textbf{30.19} & \textbf{37.38}          & 13.04          & \textbf{41.44}          & \textbf{46.26}          & 3.49          \\
EDMem-dyn.                               & 43.82          & 50.64          & 6.81          & 44.44          & 51.10          & 8.29           & 27.70          & 33.41          & 14.07          & 39.52          & 44.09          & 3.49          \\ \bottomrule
\end{tabular}}}
\caption{Exact match scores on entity answers (``Ent.'') and non-entity answers (``Non-Ent.'') in open-domain QA.}
\label{tab:linkable}
\vspace{-0.1cm}
\end{table*}

\subsubsection{Results} 

Experimental results on open-domain QA datasets are listed in Table~\ref{tab:openqa}. With the same architecture, EncDec outperforms BART due to the additional salient span masking pre-training.
Memory-based auto-encoder models like EaE and EncMem perform entity linking to provide answers. They outperform traditional encoder-decoder models by a large margin on TQA and WQ. However, target answers are mainly entities on both datasets\footnote{89\% of the answers in WQ test set are Wikipedia entities, and the ratio is 84\%/70\% for TQA/NQ.}. While on NQ where there are fewer entity answers, the performance of EncMem is similar to BART-Large.
Compared to baselines, the free-form EDMem already outperforms memory-based auto-encoder models and traditional encoder-decoder models on TQA and NQ. EDMem-static and EDMem-dynamic explicitly copy entity names into the generated answers, which further improves EDMem's performance, especially on TQA and WQ datasets where a larger portion of answers are entities. Overall, EDMem improves the best of closed-book baselines by 9\%/6\%/6\% on TQA/NQ/WQ, respectively. Although closed-book models are still behind open-book models in general, our approach shows that by combining the merits of encoder-decoder models like BART and entity linking models like EaE, the performance of EDMem is getting closer to open-book approaches and even outscores the open-book model REALM on WQ.

\subsubsection{Entity/Non-Entity Answers} 

To further investigate the improvements of EDMem over previous closed-book models, 
we calculate EM scores on two subsets divided \textit{w.r.t.} the answer type (\textit{i.e.}, entity answers and non-entity answers). 
If an answer can be directly linked to a Wikipedia entity according to Google's SLING~\cite{SLING} phrase table, it is counted as an entity answer, otherwise a non-entity answer. As shown in Table~\ref{tab:linkable}, as an entity linking model, EaE cannot predict non-entity answers; and as an encoder-decoder model, BART is able to generate a portion of non-entity answers while its accuracy on entity answers is much lower than EaE due to the lack of entity knowledge. EDMem incorporates entity knowledge into an encoder-decoder model, making it competitive in both entity answers and non-entity answers. However, the free-form generation variant is not as accurate on entity answers as EaE, because it may generate any form of answers while EaE always predicts a valid entity name. The entity linking variants, on the other hand, remedy this issue by setting constraints on generating entity names, either statically or dynamically. Such approaches improve the model performance on entity answers although sacrificing some performance on non-entity ones, and achieve the best overall performance on all answers. Besides, the best setting of EDMem outperforms EaE in entity answers as well, presumably due to a larger number of transformer layers trained in the encoder-decoder architecture, compared to its auto-encoder counterpart.

\begin{table}[t]
\setlength{\tabcolsep}{1mm}{
\resizebox{0.48\textwidth}{!}{
\begin{tabular}{cccccc}
\toprule
\textbf{Type} & \textbf{Model}              & $T_{ind}$   & $T_{ret}$ & $T_{pred}$ & \textbf{EM} \\ \midrule
\multirow{4}{*}{Closed-book} & BART  & 0  & 0       & 17s     & 25.02    \\
& EDMem-free & 0 & 0       & 28s     & 42.24
\\
& EDMem-dyn. & 0 & 0       & 48s     & 43.82
\\
& EDMem-stat. & 0 & 0       & 59s     & 46.19
\\ \midrule
Open-book & FiD     & 29min  & 15min    & 41min     & 67.60  \\ \bottomrule
\end{tabular}}}
\caption{Inference time on TriviaQA test set. $T_{ind}$ is the time for loading the index, which is a fixed amount of time; $T_{ret}$ and $T_{pred}$ denote time for document retrieval and answer prediction, which will linearly increase as the number of test examples increases. Inference time of EDMem-stat. is the accumulation of the entity linking step and the constrained generation step.}
\label{tab:efficiency}
\vspace{-0.2cm}
\end{table}

\subsubsection{Inference Efficiency}

We compare the time efficiency of different models during inference in Table~\ref{tab:efficiency}. We run EDMem, BART and the open-book model FiD on the test set of TQA (11K questions) with 8$\times$V100 GPUs. Compared to BART, EDMem needs to access a large entity memory multiple times, which slows down the inference time from 10s to 28s. However, such a time cost is much smaller than the gap between closed-book models and the open-book FiD (85min). In the open-book setting, the model needs to (1) load the pre-computed index from disk to the RAM, (2) retrieve evidence documents from the index, and (3) read all evidence documents to generate an answer. In addition to the overhead caused by accessing the index, the model needs to encode the raw text of all evidence documents (\textit{i.e.}, 100 documents for FiD) before generating an answer with the decoder, but EDMem and BART only needs to encode the question itself. Thus, EDMem is able to achieve significant improvement over traditional encoder-decoder models while retaining the efficiency advantage of closed-book models.


\begin{table*}[t]
\centering
\resizebox{0.95\textwidth}{!}{
\begin{tabular}{l|lccccccc}
\toprule
\multirow{2}{*}{\textbf{Dataset}}                                                            & \multirow{2}{*}{\textbf{Model}} & \multirow{2}{*}{\textbf{ROUGE-1}} & \multirow{2}{*}{\textbf{ROUGE-2}} & \multirow{2}{*}{\textbf{ROUGE-L}} & \multirow{2}{*}{\textbf{F1}} & \multirow{2}{*}{\textbf{BERTScore}}   & \multicolumn{2}{c}{\textbf{Entity Coverage}} \\
&&&&&&& Total & Unseen \\ \midrule
\multirow{4}{*}{MSMARCO}                                                     & BART-Large            & 56.72            & 37.62            & 53.26            & 53.86  & 89.34    & 43.87              & 22.40                     \\
                                                                             & EDMem-free     & \textbf{57.67}            & \textbf{39.45}            & \textbf{54.45}            & \textbf{55.08}     & \textbf{89.40}   & 45.53              & 25.33                     \\
                                                                             & EDMem-dyn.  & 55.96            & 37.42            & 52.84            & 52.94    & 88.57    & \textbf{51.28}              & \textbf{28.49}                    \\\cmidrule{2-9}
                                                                             & FiD (open-book)  &     60.54        &     42.96        &     57.22        &  58.12   &  90.01   &   49.63           &     32.58               \\ \midrule
\multirow{4}{*}{\begin{tabular}[c]{@{}l@{}}CREAK\\\end{tabular}} & BART-Large            & 32.87            & 14.93            & 30.34            &      30.20     & 85.68  & 46.87              & 14.98                     \\
                                                                             & EDMem-free     & \textbf{33.81}            & \textbf{16.49}            & \textbf{31.78}            &    \textbf{31.34}    & \textbf{86.15}      & 49.06              & 16.65                     \\
                                                                             & EDMem-dyn.  & 32.70            & 15.75            & 30.68            &       30.32    & 85.76    & \textbf{49.90}              & \textbf{18.76}                     \\\cmidrule{2-9}
                                                                             & FiD (open-book)  &      35.96       &     18.11        &      33.54       &  33.57   &   85.96  &     51.02           &        21.09              \\ \midrule
\multirow{4}{*}{\begin{tabular}[c]{@{}l@{}}ELI5\end{tabular}}      & BART-Large            & 25.87            & 5.89            & 22.99            &      19.38 & \textbf{83.92}       & 29.18              & 14.83                     \\
                                                                             & EDMem-free     & 27.14            & 5.70            & 23.24            &     20.19 & 83.78        & 38.76              & 18.61                    \\
                                                                             & EDMem-dyn.  & \textbf{27.48}            & \textbf{7.14}            & \textbf{23.97}            &      \textbf{20.66} & 83.58       & \textbf{45.66}              & \textbf{23.31}                     \\\cmidrule{2-9}
                                                                             & FiD (open-book)  &    25.06         &     5.98        &       22.12      &  18.48   &  83.48   &   28.06            &     16.96                \\ \midrule
\multirow{4}{*}{WoW}                                                         & BART-Large            &       19.52           &        3.42          &     \textbf{17.22}        & 15.37     &  \textbf{83.72}  & 11.78               & 4.05                       \\
                                                                             & EDMem-free     &         18.92         &    3.50              &       16.52           & 15.28     &  83.22   & 16.71              & 6.83                       \\
                                                                             & EDMem-dyn.  &          \textbf{19.54}        &      \textbf{4.00}            &     17.01             & \textbf{15.51}    &    83.23  & \textbf{17.29}             & \textbf{7.81}   \\\cmidrule{2-9}
                                                                             & FiD (open-book)  &      22.72       &      6.43       &        20.16     &  18.48   &   83.68  &   17.91            &     11.99                \\ 
                                                                             \bottomrule
\end{tabular}}
\caption{Results on entity-intensive generation datasets. \textbf{Bold} scores are best results among closed-book models.}
\label{tab:geneartion}
\end{table*}

\subsubsection{Size of Entity Memory}
\label{sec:memory_size}

\begin{figure}[t]
\centering
    \includegraphics[width=0.42\textwidth]{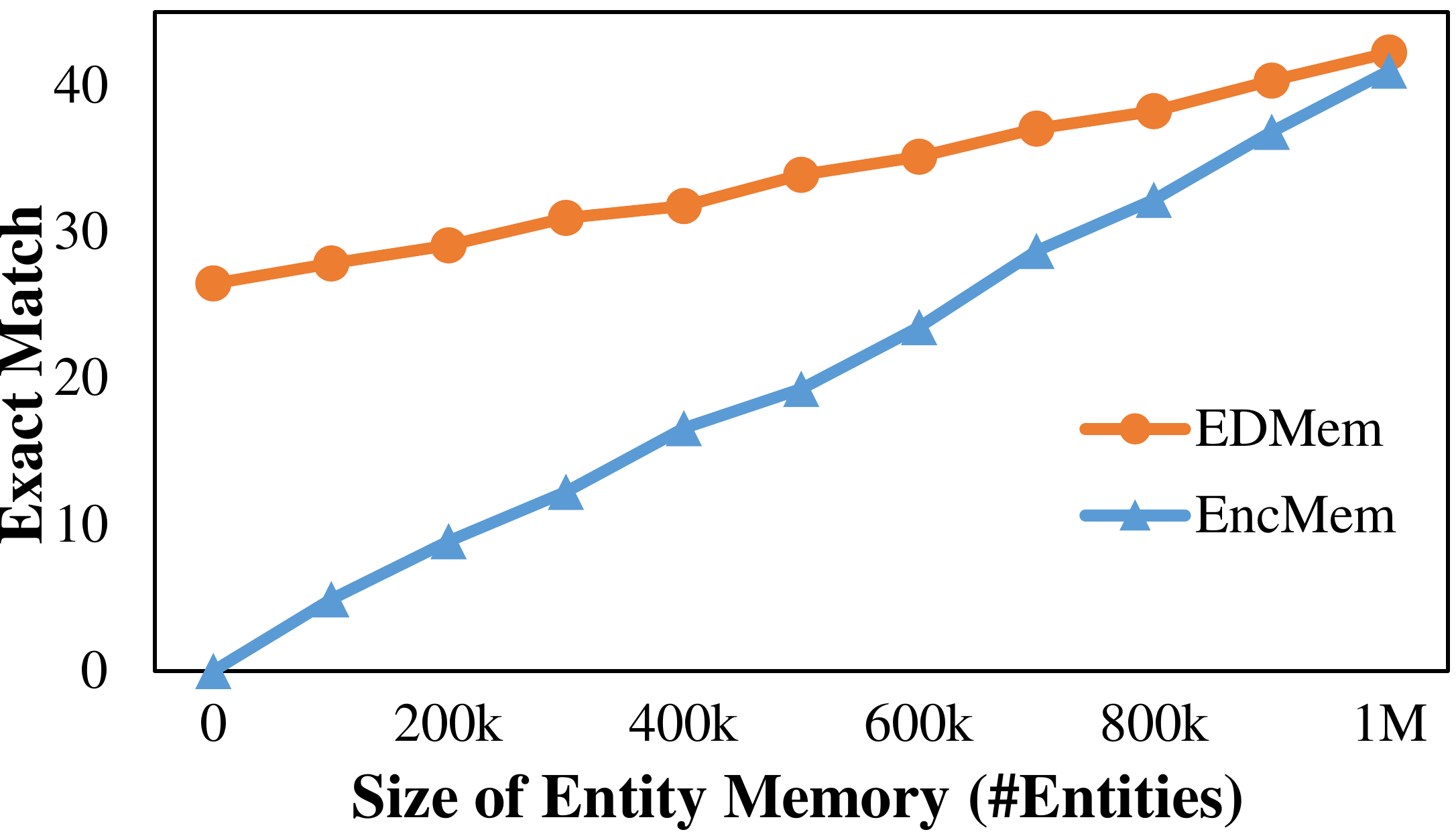}
    \caption{TQA performance of EDMem and EncMem on different memory sizes.}
    \label{fig:memory_size}
\end{figure}

We compare the performance of EDMem and its auto-encoder variant EncMem based on different sizes of the entity memory. We randomly mask out entities from the original 1M vocabulary and re-train the model. Embeddings of masked entities do not participate in computing attention while accessing the memory. According to the curves in Figure~\ref{fig:memory_size}, due to EDMem's ability of closed-book generation, it is less sensitive to the size of the entity memory, resulting in a smaller slope when less entities are visible. Particularly, EDMem is still able to generate many correct answers even when we remove the whole memory. In contrast, EncMem can only predict random entities when the entire memory is masked, which leads to a score close to zero. These results show the advantage of encoder-decoder models over auto-encoder models when jointly trained with an entity memory, especially on low-resource scenarios. 

\begin{figure}[t]
\centering
    \includegraphics[width=0.45\textwidth]{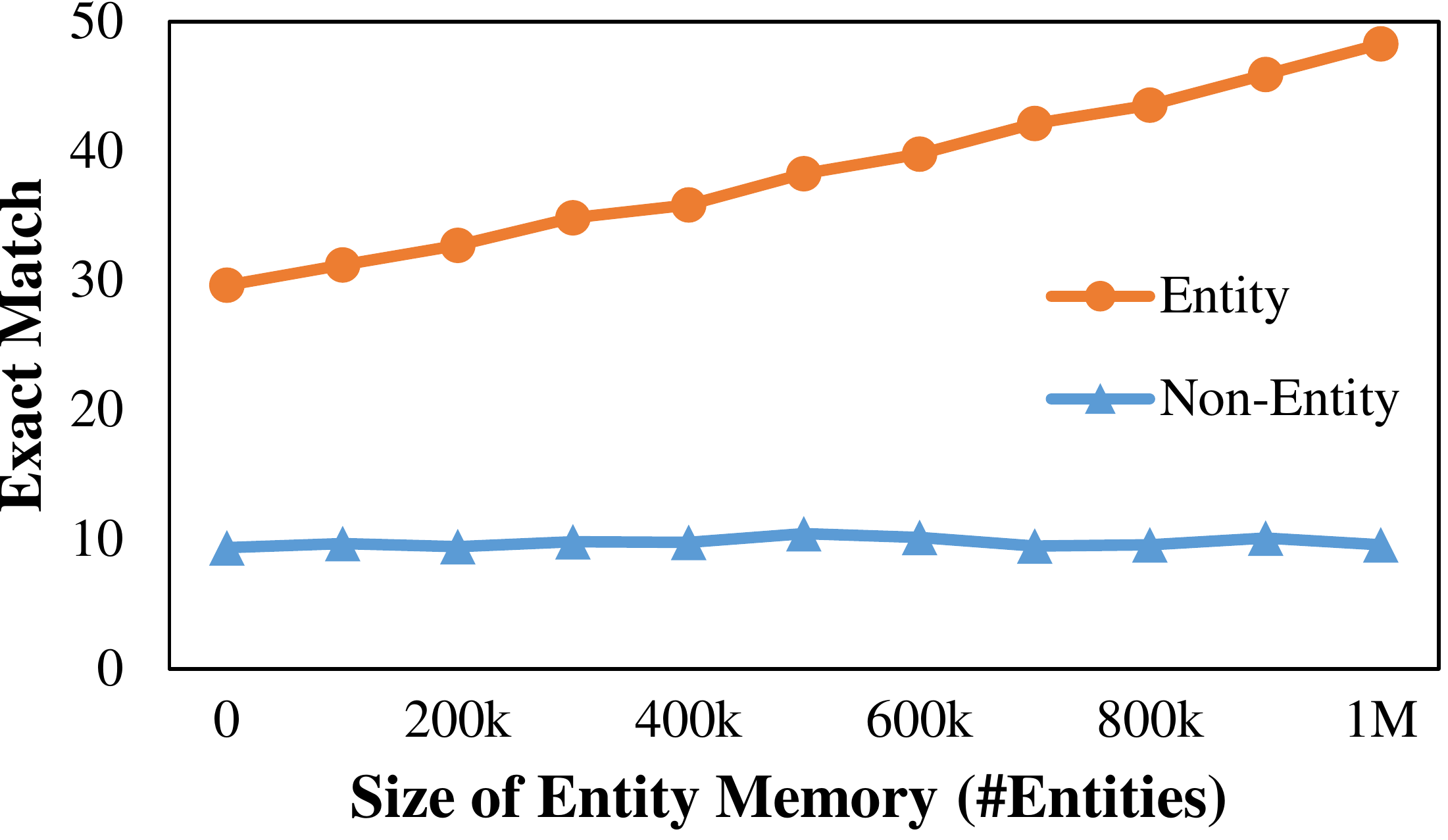}
    \caption{TQA performance of EDMem on entity and non-entity answers, trained with different memory sizes.}
    \label{fig:memory_size_linkable}
\end{figure}

In addition, we also illustrate the performance trend of EDMem on entity answers and non-entity answers in TQA. When all entities are masked, the model deteriorates to its non-memory variant EncDec. As more entity knowledge are available, EDMem performs better in predicting entity answers, while its generation performance remains consistent. These results show the advantage  of memory-based EDMem over traditional encoder-decoder models on entity-intensive task comes from the incorporation of entity knowledge from the entity memory.


\subsection{Entity-Intensive Generation}


\begin{table*}[t]
\centering
\resizebox{0.98\textwidth}{!}{\begin{tabular}{l|ccc|ccc|ccc|ccc} \toprule
\multirow{2}{*}{\textbf{Model}} & \multicolumn{3}{c|}{\textbf{MSMARCO}}       & \multicolumn{3}{c|}{\textbf{CREAK}}         & \multicolumn{3}{c|}{\textbf{ELI5}}                & \multicolumn{3}{c}{\textbf{WoW}}                 \\ 
                       & Flu.$\uparrow$ & Rel.$\uparrow$ & Cor.$\uparrow$ & Flu.$\uparrow$ & Rel.$\uparrow$ & Cor.$\uparrow$ & Flu.$\uparrow$ & Info.$\downarrow$ & Rea.$\downarrow$ & Flu.$\uparrow$ & Info.$\downarrow$ & Rea.$\downarrow$ \\ \midrule
BART-Large             &   2.90      &      1.93     &    1.57         & 2.97    & 1.82      & 1.63    &   2.77   &   2.26   &   2.34   &   2.70      &      2.45           &       2.51        \\
EDMem-free            &    2.90     &       \textbf{2.35}    &      \textbf{2.38}       & 2.99    & \textbf{2.35}     & \textbf{2.05}     &   2.86   &   \underline{1.97}    &  \underline{2.06}    &     \underline{2.83}    &      \textbf{1.86}           &      \textbf{2.14}                \\
EDMem-dynamic         &   2.87      &      \textbf{2.45}     &    \textbf{2.49}        & 2.97    & \textbf{2.32}      & \textbf{2.04}    &   2.86   &  \textbf{1.75}   &    \textbf{1.92}   &   2.80      &       \textbf{1.75}          &      \textbf{2.03}          \\ \midrule
Human         &  -      &      -     &    -   & -    & -      & -     &   2.75   &  2.01   &   2.19  &   2.68      &           1.66      &       1.86    \\ \bottomrule
\end{tabular}}
\caption{Human evaluation results. ``Flu.'', ``Rel'', ``Cor.'' stand for fluency, relevance and correctness, of which scores are given on a 1-3 scale. ``Info.'' and ``Rea.'' stand for informativeness and reasonability, of which 4 sentences are ranked \#1 -- \#4. $\uparrow$ indicates higher values are better results. $\downarrow$ indicates lower values are better results. We run paired sample \textit{t}-test comparing EDMem with BART. \textbf{Bold} scores indicate significant difference with \textit{p}-value $<$ 0.01, and \underline{underlined} scores indicate \textit{p}-value $<$ 0.05.}
\label{tab:human}
\vspace{-0.2cm}
\end{table*}

\subsubsection{Data}
\label{sec:gen_data}

To test EDMem's ability in generating longer sentences with entities, we perform experiments on several generation datasets with rich entity mentions. We choose Wizard of Wikipedia (WoW)~\cite{WoW} for knowledge-aware dialogue, MSMARCO-NLGen~\cite{MSMARCO} and ELI5~\cite{ELI5} for abstractive question answering, and CREAK~\cite{CREAK} for claim explanation\footnote{Given a claim and a true/false judgment, the model generates an explanation about why the claim is true/false.}. For MSMARCO and CREAK, we report results on the official dev set while holding out 10\% training data for validation. For ELI5, to keep a reasonable density of entities, we filter a subset of 85K data where the input and output are both no longer than 75 tokens. Detailed dataset settings are provided in Appendix~\ref{sec:appendix_dataset}.





We report ROUGE~\cite{rouge} and unigram F1 scores, as well as BERTScore~\cite{bertscore} for semantic-based evaluation. We also include metrics on evaluating entity generation. Given the entities in the ground-truth as reference, we calculate the coverage ratio of reference entities in the model-generated output. We also consider the mentions of these entities as correct matches, according to the alias table $\mathcal{T}$. 
To avoid cases where entities in the output can be directly copied from the input\footnote{For example, in CREAK dataset, the topic entity of the input claim usually appears in the explanation as well.}, we report the coverage ratio of \emph{unseen entities}, \textit{i.e.}, entities in the ground-truth output that do not exist in the input.

\subsubsection{Results}

Auto-encoder models like EaE are not applicable on these datasets, thus we compare our EDMem to the traditional encoder-decoder model BART. As shown in Table~\ref{tab:geneartion}, the free-form EDMem outperforms BART on both reference-based metrics (ROUGE, F1, BERTScore) and entity coverage scores. This indicates that the entity knowledge in the entity memory helps generate sentences with desired entities and correct entity-related information. Since these datasets cannot be directly converted to an entity linking setting, EDMem-static is not applicable here. The dynamic entity linking variant outperforms the free-form variant and BART in entity coverage scores on all datasets, while it does not sacrifice much language fluency in reference-based metrics. We find that both EDMem variants outscore BART on entity coverage by a large margin (up to 56\% on overall and up to 93\% on unseen), which indicates much stronger ability of EDMem models in entity generation. 

\subsubsection{Human Evaluation}

To test whether the model generations are reasonable to humans, we leverage Amazon's MTurk platform to conduct human evaluation. For each dataset, we sample 50 data examples with generations from BART and EDMem. We ask three annotators to evaluate each example on \textit{fluency} and two knowledge-related metrics. For CREAK and MSMARCO, knowledge-related metrics are topic \textit{relevance} and factual \textit{correctness}, given the ground-truth as reference. For ELI5 and WoW, since the ground-truth is not the only possible answer to the context, we use \textit{informativeness} and \textit{reasonability} 
as knowledge-related metrics, and we also evaluate the human-written answers. Detailed descriptions of these metrics are in Appendix~\ref{sec:appendix_human}. When evaluating \textit{informativeness} and \textit{reasonability}, annotators are asked to \textbf{rank} these generations from \#1 -- \#4, thus lower rankings indicate better results.

As shown in Table~\ref{tab:human}, EDMem generates more informative and factually correct sentences, compared to BART which lacks knowledge incorporation from the entity memory. Besides, such knowledge incorporation does not harm the fluency of model generations. In 3 out of 4 datasets, EDMem-dynamic achieves the best results on knowledge-based metrics. This indicates that integrating entity linking with text generation is beneficial for generating informative sentences with rich entity knowledge. Interestingly, annotators even prefer EDMem's generations over human answers on ELI5. One possible reason is that human answers are usually longer than model-generated ones, so not all clauses are closely related to the question. Also, the quality of some Reddit responses (\textit{i.e.}, the source of ELI5 data) may not be reliable.

\subsubsection{Case Study}

\begin{table}[t]
\centering
\resizebox{0.48\textwidth}{!}{
\begin{tabular}{p{8.15cm}}
\toprule 
\textbf{Claim}: Chicago Symphony Orchestra started in Indiana. This is false because \_\_\_\_\_ \\
\midrule
\textbf{Ground truth}: Chicago is in Illinois so it did not start in Indiana. \\ \midrule
\textbf{BART}: It was not started here. \\ \midrule
\textbf{EDMem-free}: The $[E_s]$ Chicago Symphony Orchestra $[E_e]$ started in $[E_s]$ \underline{Utah} $[E_e]$. \\
\textbf{Attended entities}: ``Illinois'', ``Chicago'', ``Cook County, Illinois'', ``Wisconsin'', ``United States''\\ \midrule
\textbf{EDMem-dynamic}: $[E_s]$ Chicago Symphony Orchestra $[E_e]$ was founded in $[E_s]$ \underline{Illinois} $[E_e]$. \\
\textbf{Attended entities}: ``Illinois'', ``Cook County, Illinois'', ``Chicago'', ``Wisconsin'', ``United States''
 \\
\bottomrule
\end{tabular}}
\caption{Case study from the CREAK dataset. We list the top-5 entities that EDMem attends to when it generates the \underline{underlined} entity.}
\label{tab:case}
\end{table}

In Table~\ref{tab:case}, we show an example from CREAK with generations of different models. Without knowledge augmentation, BART fails to generate an informative explanation on why the starting place of the orchestra is not Indiana. Although EDMem-free steps closer to the correct explanation, it falsely predicts that the orchestra started in Utah. However, ``Utah'' does not exist in the top-5 linked entities during memory access. After we constrain the generation space of EDMem-dynamic to the top-5 predicted entities, ``Utah'' is no longer valid to generate, and the model finds ``Illinois'' as the correct location. Examples from other datasets can be found in Appendix~\ref{sec:appendix_case}.

\subsubsection{Impact of Entity Richness on Generation Improvement}
In Table~\ref{tab:num_mentions}, we show detailed ROUGE-L scores according to the number of entity mentions in the ground-truth. Examples with larger numbers of mentions require more entity knowledge to generate. We list scores for CREAK dataset where the outputs are short factual claims, and ELI5 dataset where the outputs are long and diverse answers. In both datasets, the improvement of EDMem over BART occurs on entity-rich generations. For examples which do not need entity knowledge to solve (contain 0 mentions), there is not much difference between the performance of two models. This further demonstrates the effectiveness of incorporating knowledge from the entity memory on entity-intensive generation tasks.

\begin{table}[tbp]
\centering
\resizebox{0.48\textwidth}{!}{\begin{tabular}{c|cc|cc}
\toprule
\multirow{2}{*}{\textbf{\#Mentions}} & \multicolumn{2}{c|}{\textbf{CREAK}} & \multicolumn{2}{c}{\textbf{ELI5}} \\ 
                            & BART    & EDMem           & BART   & EDMem           \\\midrule
0                           & 8.89    & 8.86            & 13.75  & 14.28           \\
1                           & 26.28   & 27.47  & 15.96  & 16.38           \\
2                           & 35.33   & 37.10  & 16.41  & 17.50  \\
3                           & 34.63   & 36.55  & 17.91  & 19.76  \\
4                           & 34.05   & 34.78           & 17.14  & 19.42  \\
5+                          & 30.84   & 31.56           & 18.84  & 20.23 \\ \bottomrule
\end{tabular}}
\caption{ROUGE-L scores based on different number of mentions in the ground-truth reference.}
\label{tab:num_mentions}
\end{table}

\section{Conclusions}
\label{sec:conclusions}
In this work, we proposed EDMem, an encoder-decoder framework with entity memory. The entity memory was pre-trained on Wikipedia to provide entity knowledge for the encoder-decoder model. EDMem also performed entity linking with the memory to assist entity generation in downstream tasks. As a unified framework, EDMem outperformed previous closed-book models on various entity-intensive QA and generation tasks, and still retained the efficiency advantage over open-book models.  Further analysis proved that the proposed EDMem was enabled for entity linking with the entity memory and for generation with the encoder-decoder framework.


\section{Limitations}
\label{sec:limitation}
First, if applying EDMem to other datasets, its performance may correlate to the density of entity mentions in data examples. EDMem may not be able to acquire sufficient entity knowledge from the memory if there are few mentions in the specific task. Another limitation of our work is that the pre-trained entity memory may not be generalized to special domains, \textit{e.g.}, biomedical text. A lot of specific terminology is not included in our pre-trained entity memory, which may require additional training on domain-specific corpora.

\section*{Acknowledgement}
\label{sec:acknowledgement}
This work was supported in part by NSF IIS-1849816, IIS-2119531, IIS-2137396, IIS-2142827, CCF-1901059, and ONR N00014-22-1-2507. We would like to thank Yuwei Fang (Microsoft), Jinfeng Lin (Meta), Mingxuan Ju (University of Notre Dame), and Qian Liu (Nanyang Technology University) for their valuable suggestions to this work.

\nocite{koala}
\nocite{multi-task}
\nocite{dca}
\nocite{genread}

\balance
\bibliography{reference}
\bibliographystyle{acl_natbib}

\clearpage
\appendix
\section{Pre-Training}
\label{sec:appendix_pretrain}
\subsection{Pre-Training Data}
We pre-train our model on the Wikipedia corpus of over 5 million documents. All documents are split into 128-token passages. The last passage is round up to 128 tokens by appending tokens from the beginning of the same document, so there are no cross-document passages. In addition, we set a 10-token sliding window between passages to avoid an entity being split into two adjacent chunks. Such a setting yields a total of 39 million passages, of which we hold out 0.5\% of them as the validation set during the pre-training process. For supervision signals on the entity memory, we leverage Wikipedia hyperlinks as gold annotations. Each hyperlink provides the boundaries of a mention, and also the corresponding entity\footnote{In Wikipedia, each page corresponds to a unique entity which has the same name as the page title.} that the mention is linked to. 

However, the average density of Wikipedia hyperlinks is only one in 21 words, which means 6 mentions per passage. This is because in a specific page, (1) only the first mention of an entity is linked and (2) the title entity is not linked since a hyperlink always redirects to a different page. To provide more supervision signals for entity embedding learning, we label the missing mentions using heuristic rules. We use the alias table $\mathcal{T}$ to label all mentions in a Wikipedia page if they match either (1) a linked entity in the same page, or (2) the title entity of this page. After such heuristic labeling, the hyperlink density increases to one in 11 words, with a passage having 12 mentions on average. We manually checked 50 passages and found the precision of such heuristic labeling to be 92\%, a pretty acceptable rate.

\subsection{Pre-Training Settings}

We pre-train our model on the Wikipedia corpus containing 39 million passages for 1 million steps using a batch size of 2048. AdamW~\cite{AdamW} optimizer is used with maximal learning rate $1\times 10^{-4}$ and the weight decay coefficient is 0.01. The learning rate is scheduled to be warmed up for 10\% of the training steps and then linearly decay. The mask rate for random token masking is $P_{rtm}=0.3$, and the mask rate for salient span masking is $P_{ssm}=0.5$ (ablations in Appendix \ref{sec:mask_rate}). The maximum length of input sequence is set to 128. The coefficient of the entity linking objective is set to $\lambda_{EL}=1.0$ and the dropout rate is 0.1. The whole model is trained from scratch. We tried to initialize the encoder-decoder model with BART~\cite{BART} and derive entity embeddings from BART embeddings, but the model showed up to be unstable in further training. We use the mixed precision floating point arithmetic~\cite{fp16} to speed-up training. The full setting of EDMem takes about two weeks to train on 16$\times$A100 GPUs.

\section{Fine-Tuning}

Different from pre-training on Wikipedia, in open-domain QA and generation tasks, there are no gold annotations of mention boundaries in the input and output. Therefore, we annotate mention boundaries as well as the linked entities using a state-of-the-art neural entity linker ELQ~\cite{ELQ}. For generation datasets, we pass the source sequence and the target sequence into the ELQ model respectively, and obtain their mention annotations. For open-domain QA datasets, since the answers are usually short, we concatenate the question and the answer as input to the ELQ model.
During fine-tuning, we tune the hyperparameters within the following ranges: learning rate  $\in\{\text{5e-6},\text{1e-5},\text{2e-5},\text{3e-5}\}$, $\lambda_{EL} \in\{0.5,1.0,2.0\}$, dropout rate $\in\{0.1,0.2,0.3\}$, beam size $\in\{1,3,5\}$, \#candidate entities (in static/dynamic entity linking) $\in\{1,3,5\}$. They are tuned based on on the main evaluation metric of the specific task (open-domain QA: EM; WoW: F1; other generation datasets: ROUGE-L) on the dev set. Batch size is fixed to 256 unless it exceeds GPU memory or the dataset is too small (\textit{e.g.}, CREAK and WQ). Early stopping is used with 20 waiting steps on the dev set. Fine-tuning EDMem usually costs a few hours (\textit{e.g.}, \textasciitilde3 hours on TQA) on 8$\times$V100 GPUs.

\section{Entity Memory Settings}
We collect the 1-million most frequent entities in Wikipedia documents as our entity vocabulary $\mathcal{E}$. The frequency of an entity is calculated based on how many hyperlinks are linked to the Wikipedia page of that entity. The dimension of entity embeddings learned in the memory is set to 256. 
The model attends to all 1 million entities during training. During inference, top-100 entities are selected according to the dot product similarity, and we only integrate the embeddings of these 100 entities when performing attention.


\section{Datasets}

\subsection{Open-Domain QA Datasets}
\begin{table}[t]
\centering
\resizebox{0.43\textwidth}{!}{\begin{tabular}{l|ccc}
\toprule
\textbf{Dataset} & \textbf{Train} & \textbf{Dev} & \textbf{Test} \\ \midrule
TQA (In-House)   & 78,785          & 8,837         & 11,313         \\
TQA (Official)   & 87,622          & 11,313        & Not Used      \\
NQ               & 79,168          & 8,757         & 3,610          \\
WQ               & 3,417           & 361          & 2,032         \\ \bottomrule
\end{tabular}}
\caption{Statistics of open-domain QA datasets.}
\label{tab:qa_dataset}
\end{table}
\begin{table}[t]
\centering
\resizebox{0.37\textwidth}{!}{\begin{tabular}{l|ccc}
\toprule
\textbf{Dataset} & \textbf{Train} & \textbf{Dev} & \textbf{Test} \\ \midrule
MSMARCO          & 138,352         & 15,373        & 12,467         \\
CREAK            & 9,158           & 1,018         & 1,371          \\
ELI5             & 75,220          & 8,361         & 1,384          \\
WoW              & 54,330          & 3,054         & 2,944         \\ \bottomrule
\end{tabular}}
\caption{Statistics of generation datasets.}
\label{tab:gen_dataset}
\end{table}

Statistics of the open-domain QA datasets are listed in Table~\ref{tab:qa_dataset}. In TQA, most previous works used the in-house split provided by~\cite{lee2019latent}, while we also test EDMem on the official dev set to compare with the scores reported by EaE.

\subsection{Generation Datasets}
\label{sec:appendix_dataset}
Here we provide detailed descriptions of the generation datasets used in our experiments. Statistics of these datasets are listed in Table~\ref{tab:gen_dataset}.

\paragraph{MSMARCO} MSMARCO~\cite{MSMARCO} is originally collected for the abstractive QA task. We use the NLGen split where answers are sentences carefully written by human workers. Since the official leaderboard has been closed, we test our model on the official dev set and hold out 10\% training examples for validation.

\paragraph{CREAK} CREAK~\cite{CREAK} is a recent dataset for claim verification and explanation. In our experiments, we target on the explanation sub-task. The model is given a factual claim and a true/false judgment, and is expected to generate an explanation about why the claim is true or false. Since the official test set does not have explanations, we report results on its dev set and hold out 10\% of the training set for validation.

\paragraph{ELI5} ELI5~\cite{ELI5} is a dataset for generating long-form responses for factual questions. To keep a reasonable density of entities, we filter a subset where the input question and the output response are both no longer than 75 tokens. We also remove those examples which have no entity mentions in the output. This result in a total of 85K data examples.

\paragraph{Wizard of Wikipedia (WoW)} WoW~\cite{WoW} is a dialogue dataset where entity knowledge is included in speakers' responses. We use the open-domain setting of this dataset provided by the KILT benckmark~\cite{KILT}, where no candidate knowledge pieces are given to make the response. We additionally remove the training examples where no knowledge piece is used to generate the response.






\section{Additional Experiments}

\subsection{Pre-Training Mask Rates}
\label{sec:mask_rate}

We test the performance of EDMem on different mask rates during pre-training, and list the results in Table~\ref{tab:mask}. In pre-training, we adopt two masked language modeling objectives: random token masking (RTM) and salient span masking (SSM). When using smaller mask rates, there is more visible contextual information to the model. Therefore, when evaluating the model on the validation set during pre-training, smaller mask rates lead to lower language model perplexity and better entity linking performances.
However, larger mask rates finally lead to better performances on the downstream task. With larger mask rates, more training signals are applied to the model and thus the model is more sufficiently trained. Specifically, in SSM, the model is encouraged to leverage the entity knowledge from the entity memory to predict the masked mention. Therefore, a larger SSM rate leads to more sufficient learning of the entity memory, where the contextual information of masked mentions is integrated into the corresponding entity embedding.

\subsection{Pre-Training Steps}

\begin{figure}[t]
    \centering
    \includegraphics[width=0.45\textwidth]{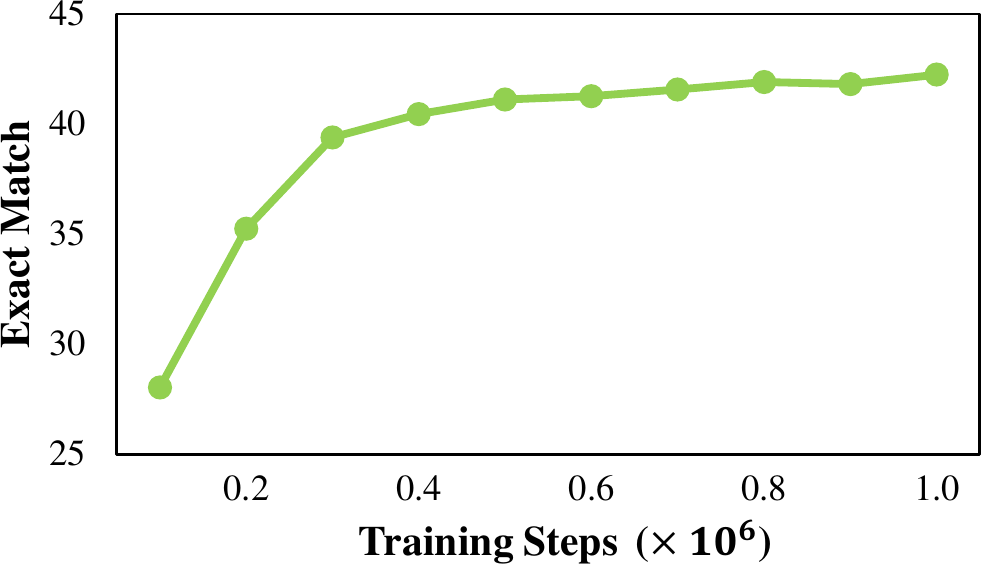}
    \caption{TQA performance of EDMem with different pre-training steps.}
    \label{fig:training_step}
\end{figure}

We fine-tune EDMem on TQA using pre-trained checkpoints of different number of training steps. As is shown in Figure~\ref{fig:training_step}, longer pre-training leads to better performance on the downstream task. Although there is no sign of overfitting, as the learning rate gradually decays and the model gradually converges, there is not much improvement of the model performance after 500K steps.

\begin{table}[t]
\centering
\resizebox{0.4\textwidth}{!}{\begin{tabular}{cc|cc|cc}
\toprule
\multirow{2}{*}{$P_{rtm}$} & \multirow{2}{*}{$P_{ssm}$} & \multicolumn{2}{c|}{\textbf{Pre-Train}} & \multicolumn{2}{c}{\textbf{TQA}} \\
                                &                                 & PPL$\downarrow$               & ACC$\uparrow$                & EM$\uparrow$             & Entity$\uparrow$          \\ \midrule
0.1                             & 0.3                            & 1.14              & 73.33              & 41.01           & 46.86          \\
0.2                             & 0.4                             & 1.31              & 72.94              & 41.52           & 47.03          \\
0.3                             & 0.5                             & 1.51              & 71.37              & 42.24           & 48.27               \\ \bottomrule
\end{tabular}}
\caption{Performance of EDMem with different pre-training mask rates. PPL: perplexity of the masked tokens on validation set (lower is better); ACC: entity linking accuracy of masked mention spans on validation set; EM: overall exact match scores; Entity: exact match scores on entity answers.}
\label{tab:mask}
\end{table}


\section{Human Evaluation Details}
\label{sec:appendix_human}

\begin{table*}[bp]
\centering
\resizebox{0.9\textwidth}{!}{
\begin{tabular} {p{16.15cm}}
\toprule 
\textbf{Question}: At the \underline{equator}, in \underline{miles per hour}, what speed of the \underline{ground} beneath your feet, as a result of the \underline{Earth}'s \underline{rotation}? \\
\midrule
\textbf{Ground truth}: [18,000 mph, eighteen thousand speed, 18,000 speed] \\ \midrule
\textbf{BART}: 8,000 \quad \textbf{EncMem}: Speed \quad \textbf{EDMem-free}: 18,000 mph
 \quad \textbf{EDMem-static}: 18,000 speed
 \\
\bottomrule
\end{tabular}}
\caption{Case study of an example with a non-entity answer from the TQA dataset. Entity mentions in the question are \underline{underlined}. Since the auto-encoder model EncMem can only perform entity linking to provide an answer, it selects an entity ``Speed'' from the entity memory. In contrast, EDMem has the ability of generating non-entity answers thanks to the encoder-decoder framework, and successfully predicts the correct amount of speed.}
\label{tab:case_tqa}
\end{table*}

\begin{table*}[bp]
\centering
\resizebox{0.93\textwidth}{!}{
\begin{tabular}{p{17.15cm}}
\toprule 
\textbf{Question}: Where is Niagara? \\
\midrule
\textbf{Ground truth}: Niagara is in New York, United States. \\ \midrule
\textbf{BART}: Niagara is in Ontario, Canada. \\ \midrule
\textbf{EDMem-free}: $[E_s]$ Niagara $[E_e]$ is in $[E_s]$ \underline{Niagara County, Pennsylvania} $[E_e]$, $[E_s]$ United States $[E_e]$. \\
\textbf{Attended entities}: ``Niagara County, New York'', ``Regional Municipality of Niagara'', ``Niagara Falls, New York'', ``Niagara Falls'', ``Wayne County, New York''\\ \midrule
\textbf{EDMem-dynamic}: $[E_s]$ Niagara Falls $[E_e]$ is in $[E_s]$ \underline{Niagara County, New York} $[E_e]$, $[E_s]$ United States $[E_e]$. \\
\textbf{Attended entities}: ``Niagara County, New York'', ``Niagara Falls'', ``Regional Municipality of Niagara'', ``New York (state)'', ``Upstate New York''
 \\
\bottomrule
\end{tabular}}
\caption{Case study from the MSMARCO dataset. We list the top-5 entities that EDMem attends to when it generates the \underline{underlined} entity. EDMem-free correctly predicts the county name and country name of Niagara. However, the state name ``Pennsylvania'' is not expected. Although the model generates $[E_s]$ and $[E_e]$ tokens around ``Niagara County, Pennsylvania'', it is not a valid entity name which leads to false information. In contrast, EDMem-dynamic generates the correct answer by constraining the model to generate ``Niagara County, New York'', one of the top-5 selected entity names.}
\label{tab:case_msmarco}
\end{table*}

\begin{table*}[bp]
\centering
\resizebox{0.9\textwidth}{!}{
\begin{tabular}{p{16.15cm}}
\toprule 
\textbf{Question}: The history of Canada \\
\midrule
\textbf{Human-Written Answer}: Canada was initially settled by France and later Britain invaded. Later some of the Southern British colonies rebelled against the King but the Northern colonies chose not to. A hundred years after that Southern rebellion the Northern colonies were granted their independence peacefully. \\ \midrule
\textbf{BART}: Canada is a country of about 5 million people. It's a very large country, and it's been a part of the world for a long time. \\ \midrule
\textbf{EDMem-free}: $[E_s]$ Canada $[E_e]$ is a $[E_s]$ country $[E_e]$ in the $[E_s]$ North America $[E_e]$. It was a $[E_s]$ colony $[E_e]$ of $[E_s]$ \underline{English} $[E_e]$ and $[E_s]$ French $[E_e]$. \\
\textbf{Attended entities}: ``Great Britain'', ``United Kingdom'', ``England'', ``British Empire'', ``British people''\\ \midrule
\textbf{EDMem-dynamic}: $[E_s]$ Canada $[E_e]$ is a country in the $[E_s]$ Americas $[E_e]$. It was a part of the $[E_s]$ \underline{British Empire} $[E_e]$, which was a $[E_s]$ nation $[E_e]$ of $[E_s]$ Europe $[E_e]$. \\
\textbf{Attended entities}: ``British Empire'', ``Commonwealth of Nations'', ``United Kingdom'', ``Canada'', ``United States''
 \\
\bottomrule
\end{tabular}}
\caption{Case study from the ELI5 dataset. We list the top-5 entities that EDMem attends to when it generates the \underline{underlined} entity. BART generates a fluent sentence about Canada. However, it barely contains any useful information. Canada's population is much larger than 5 million, and the whole generation is not related to the history of Canada. The generation of EDMem-free is more informative than BART, but when it is supposed to generate country names, it generates ``English'' and ``French'', which hurts its performance. EDMem-dynamic avoids this problem since ``English'' is not in the top attended entities. It is able to generate a more informative answer by correctly selecting entities from the memory.}
\label{tab:case_eli5}
\end{table*}

\begin{table*}[htbp]
\centering
\resizebox{0.9\textwidth}{!}{
\begin{tabular}{p{16.15cm}}
\toprule 
\textbf{Dialogue}: -- I was a cheerleader in high school. Were you ever a cheer leader? \\
-- Yeah! Remember back then, we do encourage our college basketball team to success. \\
-- What is your favorite cheer leading move? I loved the pyramid. \\
\midrule
\textbf{Human-Written Response}: Yeah I loved the components of tumbling, dance, jumps, cheers, and stunting. \\ \midrule
\textbf{BART}: I love the pyramid. It is a large, circular structure that is used to hold the flag. \\ \midrule
\textbf{EDMem-free}: I like the $[E_s]$ \underline{hula hoop} $[E_e]$. It is a $[E_s]$ dance $[E_e]$ performed by $[E_s]$ hula hoops $[E_e]$. \\
\textbf{Attended entities}: ``Haka'', ``Human pyramid'', ``Jumping'', ``Push-up'', ``Tumbling (gymnastics)''\\ \midrule
\textbf{EDMem-dynamic}: I love the $[E_s]$ \underline{human pyramid} $[E_e]$. $[E_s]$ Cheerleading $[E_e]$ is a form of $[E_s]$ performance art $[E_e]$ that combines $[E_s]$ gymnastics $[E_e]$ and $[E_s]$ dance $[E_e]$. \\
\textbf{Attended entities}: ``Human pyramid'', ``Jumping'', ``Haka'', ``Cheering'', ``Cheerleading''
 \\
\bottomrule
\end{tabular}}
\caption{Case study from the WoW dataset. We list the top-5 entities that EDMem attends to when it generates the \underline{underlined} entity. The second generated sentence of BART does not correlate with the context, and does not contain any correct information either. The generation of EDMem-free talks about the hula hoop, but it is not a typical cheerleading move. EDMem-dynamic generates a reasonable response by first selecting a correct cheerleading move ``human pyramid'', and then continues to generate an informative sentence by selecting relevant entities from the memory.}
\label{tab:case_wow}
\end{table*}

Here we provide the actual questions that we asked Amazon Mturk annotators in human evaluation, along with their rubrics. For \textbf{fluency}, \textbf{relevance} and \textbf{correctness} metrics, annotators are asked to give scores on a 1-3 scale. For \textbf{informativeness} and \textbf{reasonability}, ranking evaluation is applied. This is because in ELI5 and WoW datasets, the human-written answer is not the only possible one to the context, so we do not compare model generations to the ground-truth. Instead, we let annotators evaluate the human written answer along with the model-generated ones. Since \textbf{informativeness} and \textbf{reasonability} are hard to set clear rubrics if no ground-truth is given, we adopt a ranking-based evaluation. The annotator is asked to rank all sequences (three model generations and the human-written answer) from \#1 -- \#4, with lower rankings indicating better results.

\begin{itemize}
    \item \textbf{Fluency}: How is the fluency of the machine-generated explanation? (Do not consider its correctness)
    \begin{itemize}
        \item[$\diamond$] 3 -- Fluent English
        \item[$\diamond$] 2 -- Readable with grammar errors or typos
        \item[$\diamond$] 1 -- Not fluent at all
    \end{itemize}
    \item \textbf{Relevance}: Does the machine-generated explanation contain the same concepts as the human-written reference? (Synonyms are allowed, and do not consider its factual correcteness)
    \begin{itemize}
        \item[$\diamond$] 3 -- Contains the same concepts as the reference
        \item[$\diamond$] 2 -- Misses some concepts in the reference, or contains redundant concepts
        \item[$\diamond$] 1 -- Does not contain any concept in the reference
    \end{itemize}
    \item \textbf{Correctness}: Does the machine-generated explanation express similar meanings with the human-written reference? (Paraphrases are allowed)
    \begin{itemize}
        \item[$\diamond$] 3 -- Expresses similar meanings with the reference
        \item[$\diamond$] 2 -- Expresses partial meanings of the reference
        \item[$\diamond$] 1 -- Expresses totally different meanings with the reference
    \end{itemize}
    \item \textbf{Informativeness}: Rank these answers based on the amount of information they contain. \#1: most informative, \#4: least informative. Ties are allowed (e.g., 1/1/3/4 or 1/2/2/2). You do not need to consider whether the information is relevant to the question.
    \item \textbf{Reasonability}: Rank these answers based on whether they are reasonable answers to the question. \#1: most reasonable, \#4: least reasonable. Ties are allowed (e.g., 1/1/3/4 or 1/2/2/2).
\end{itemize}

\section{Case Study}
\label{sec:appendix_case}

We present an example on TQA with a non-entity answer in Table~\ref{tab:case_tqa}. We use our auto-encoder variant EncMem to represent memory-based auto-encoder models. We also provide additional examples on generation datasets in Tables~\ref{tab:case_msmarco} --~\ref{tab:case_wow}.

\end{document}